\documentclass{article}
\usepackage{spconf}
\usepackage{amsmath,graphicx}
\usepackage{regensky}
\usepackage[firstpage=true]{background}
\usepackage{hyperref}

\SetBgContents{\parbox{\textwidth}{\footnotesize © 2021 IEEE. Personal use of this material is permitted. Permission from IEEE must be
obtained for all other uses, in any current or future media, including
reprinting/republishing this material for advertising or promotional purposes, creating new
collective works, for resale or redistribution to servers or lists, or reuse of any copyrighted
component of this work in other works. DOI: \href{https://doi.org/10.1109/ICASSP39728.2021.9413576}{10.1109/ICASSP39728.2021.9413576.}}}
\SetBgScale{1}
\SetBgAngle{0}
\SetBgPosition{current page.north}
\SetBgVshift{-1.6cm}
\SetBgColor{black}
\SetBgOpacity{1}

\title{A NOVEL VIEWPORT-ADAPTIVE MOTION COMPENSATION TECHNIQUE\\FOR FISHEYE VIDEO}

\name{Andy Regensky, Christian Herglotz, and André Kaup\thanks{The authors gratefully acknowledge that this work has been supported by the German Research Foundation (DFG) under contract number KA 926/9-1.}}
\address{Multimedia Communications and Signal Processing\\Friedrich-Alexander University Erlangen-Nürnberg (FAU), Cauerstr. 7, 91058 Erlangen, Germany}

\begin{document}

  \ninept
\maketitle
\begin{abstract}
Although fisheye cameras are in high demand in many application areas due to their large field of view, many image and video signal processing tasks such as motion compensation suffer from the introduced strong radial distortions. A recently proposed projection-based approach takes the fisheye projection into account to improve fisheye motion compensation. However, the approach does not consider the large field of view of fisheye lenses that requires the consideration of different motion planes in 3D space. We propose a novel viewport-adaptive motion compensation technique that applies the motion vectors in different perspective viewports in order to realize these motion planes. Thereby, some pixels are mapped to so-called virtual image planes and require special treatment to obtain reliable mappings between the perspective viewports and the original fisheye image. While the state-of-the-art ultra wide-angle compensation is sufficiently accurate, we propose
a virtual image plane compensation that leads to perfect mappings. All in all, we achieve average gains \mbox{of +2.40 dB} in terms of PSNR compared to the state of the art in fisheye motion compensation.
\end{abstract}

\section{Introduction}
\label{sec:introduction}

In recent years, fisheye lenses \cite{Miyamoto1964} have found their way into many modern imaging products and are not only used for industrial applications like autonomous driving \cite{Cui2019} and video surveillance \cite{Findeisen2013}, but are also increasingly being built into consumer devices like action cameras and smartphones. The popularity of fisheye lenses stems from their large fields of view (FOV) of up to 180° and more, which enables them to capture exceptionally large regions of their surroundings. Estimating the motion between two frames of an image sequence is a crucial part of many image and video signal processing tasks like video coding \cite{Sullivan2012}, \cite{Bross2020}, framerate up-conversion \cite{Byung-TaeChoi2000}, or resolution enhancement \cite{SungCheolPark2003}. Commonly, block matching \cite{ShanZhu2000} incorporating a translational motion model is used, where a block in the current frame is matched to a block in a reference frame by shifting it according to different motion vector candidates and selecting the best match. The motion between two frames can then be compensated by copying the motion compensated blocks from the reference frame to the the corresponding blocks in a compensated frame.

Unlike traditional perspective lenses, fisheye lenses do not follow the pinhole model. They follow projection functions \cite{Miyamoto1964}, \cite{Kannala2006} that heavily bend the incident light rays towards the image sensor to realize their large FOVs. The hereby introduced strong radial distortions in the image impair the performance of block-based motion compensation techniques. There have been different approaches to improve fisheye motion compensation. Jin et al. \cite{Jin2015} proposed a pixel-wise warping of the motion vector according to an equidistant fisheye projection. Ahmmed et al. \cite{Ahmmed2016} inserted an additional reference frame generated using an elastic motion model into the motion compensation procedure. Eichenseer et al. \cite{Eichenseer2019} introduced a state-of-the-art projection-based approach where the block matching procedure is conducted in the perspective domain.

\begin{figure}[t]
\begin{minipage}[b]{0.3\columnwidth}
  \centering
  \centerline{\includegraphics[width=\linewidth]{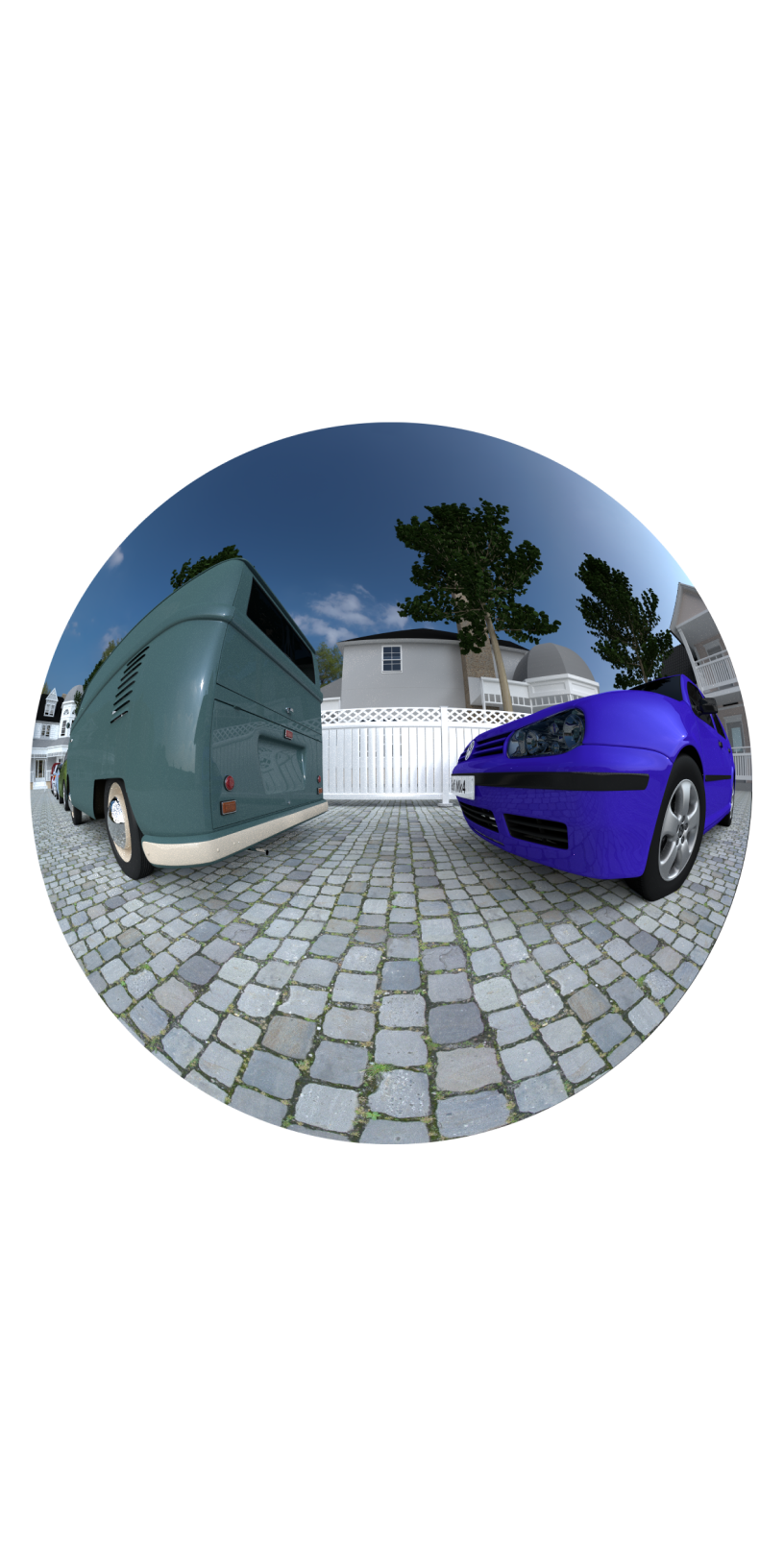}}
  \centerline{a) Fisheye image.}\medskip
\end{minipage}
\hfill
\begin{minipage}[b]{0.6\columnwidth}
  \centering
  \centerline{\includegraphics[width=\linewidth]{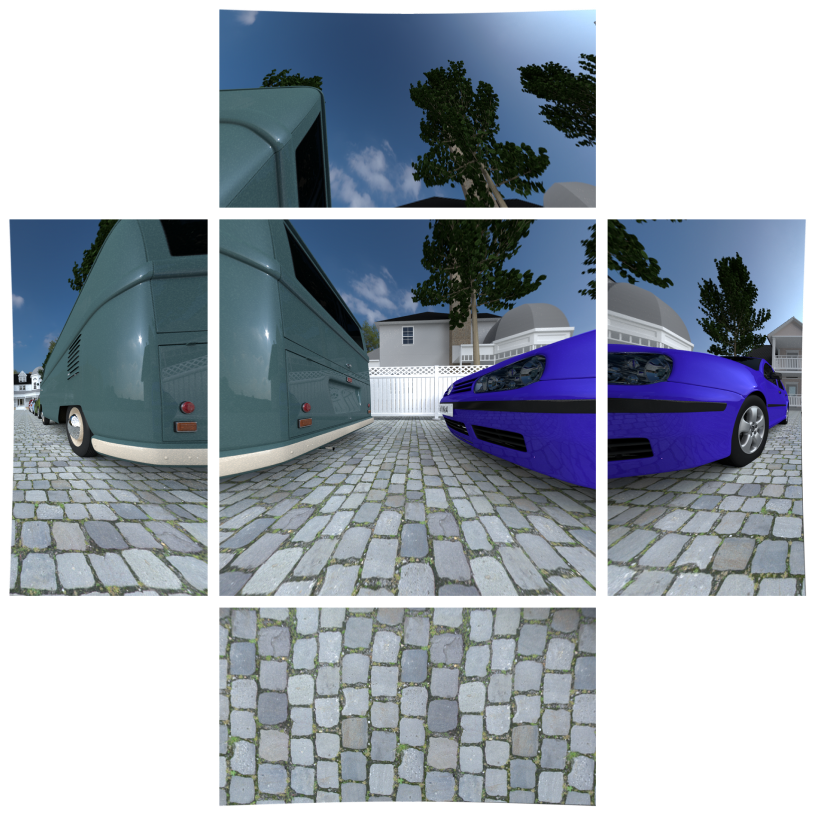}}
  \centerline{b) Perspective viewports.}\medskip
\end{minipage}
\caption{a) The original fisheye image \cite{Eichenseer2016} and b) the top, left, front, right and bottom perspective viewports.}
\label{image:viewport-introduction}
\end{figure}

Although the latter projection-based approach takes the fisheye projection into account during motion compensation, it does not cover for the large incident angles of up to 90° and more, where different motion planes need to be considered. Thereby, a motion plane describes a plane in 3D space on which the motion is performed according to a translational model. Implicitly, the traditional approach as well as the projection-based approach take only the forward oriented motion plane into account, i.e.\ the motion plane parallel to the sensor plane. While this is sufficient for perspective lenses, the large FOV of fisheye lenses requires different motion planes to be considered. Similar to 360° images, fisheye images can be mapped to a sphere in 3D space. Hence, generating different viewports into fisheye images is possible as shown in Fig. \ref{image:viewport-introduction}b) where the top, left, front, right and bottom viewports into the original fisheye image in Fig. \ref{image:viewport-introduction}a) are visualized. In the novel viewport-adaptive approach, we exploit this possiblity of projecting the fisheye coordinates to different perspective viewports in order to realize differently oriented motion planes in 3D space.

The remainder of this paper is organized as follows: Section \ref{sec:ptmc} briefly summarizes the projection-based approach from \cite{Eichenseer2019}. Section \ref{sec:mv-ptmc} explains the novel viewport-adaptive approach in detail and introduces an improved virtual image plane compensation. Section \ref{sec:experiments} evaluates the performance of the new approach on the task of motion compensation and gives an estimate of the expected rate-distortion performance in video coding compared to traditional block matching. Finally, Section \ref{sec:conclusion} concludes this paper.

\section{Motion Compensation For Fisheye Video}
\label{sec:ptmc}

In traditional block matching, a motion compensated block $\boldsymbol{\hat{B}}_\text{cur}$ for the current block $\boldsymbol{B}_\text{cur}$ is obtained from a reference frame $\boldsymbol{I}_\text{ref}$ by shifting the regarded block in the reference frame according to a given motion vector candidate $\boldsymbol{m} = (\Delta x, \Delta y)$. The motion vector that leads to the minimum residual error, e.g., sum of squared differences (SSD), between $\boldsymbol{\hat{B}}_\text{cur}$ and $\boldsymbol{B}_\text{cur}$ is then selected. An entire motion compensated frame $\boldsymbol{\hat{I}}_\text{cur}$ can be obtained by taking over the best matches for all blocks in the image. While this approach provides good motion compensation quality for perspective lenses, the quality decreases notably, when fisheye lenses are applied. Thereby, the perspective projection is defined as
\begin{align}
    r_\text{p}(\theta) = f \cdot \tan(\theta) \label{eq:perspective}
\end{align}
with the focal length $f$, the incident angle $\theta$ and the radius on the image plane $r_p$. While there exist four typical fisheye projectons \cite{Miyamoto1964}, \cite{Kannala2006}, we restrict our investigations to the equisolid projection without loss of generality
\begin{align}
    r_\text{f}(\theta) = 2f \cdot \sin(\theta/2). \label{eq:fisheye}
\end{align}

The projection-based approach from \cite{Eichenseer2019} adapts the block matching procedure according to the fisheye projection as follows. While the overall procedure remains the same, the motion vectors are interpreted in the perspective domain instead of the fisheye domain directly. This is done using a so-called reprojection, where the pixel coordinates of the regarded block are projected into the perspective domain, first. The motion vector is then added to the pixel coordinates in the perspective domain and the obtained motion compensated coordinates are projected back to the fisheye domain. The pixel values corresponding to the motion compensated coordinates can then be extracted from the original fisheye image. Note that an interpolation of the original fisheye image is necessary, as the resulting pixel values do not necessarily lie on the integer grid.
\section{Viewport-Adaptive Motion Compensation}
\label{sec:mv-ptmc}

\begin{figure}[t]
\centering
\includegraphics[width=\columnwidth]{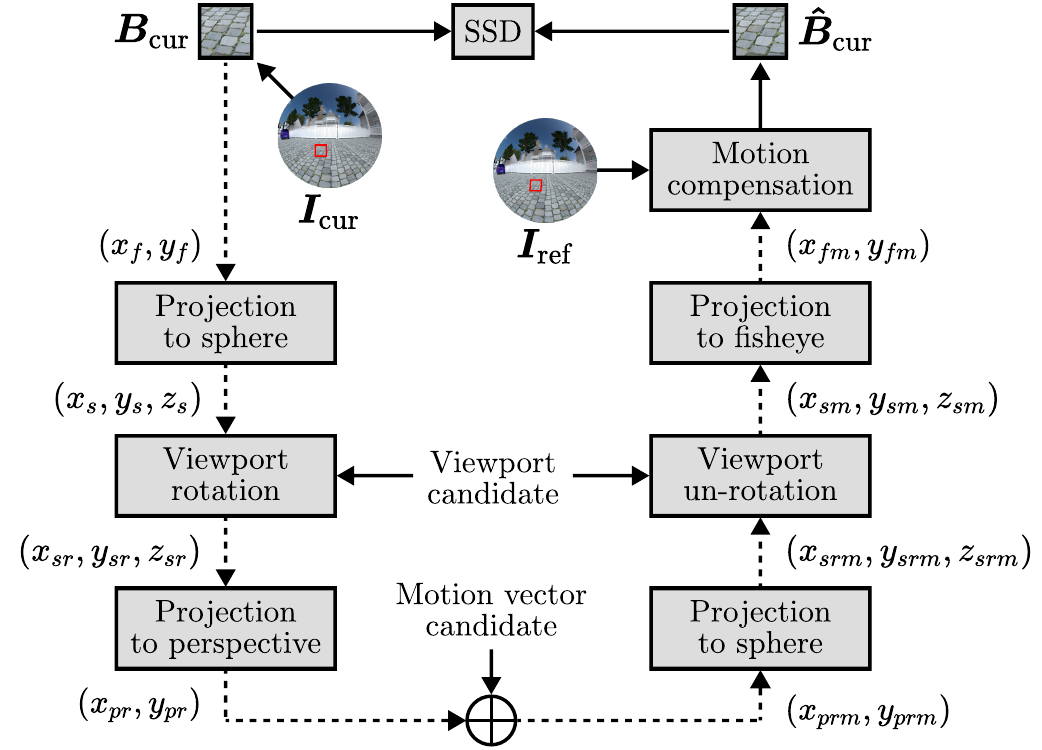}
\caption{Block diagram of the described viewport-adaptive motion compensation technique for one image block $\boldsymbol{B}_\text{cur}$ (red box), viewport candidate and motion vector candidate. Dashed lines convey pixel coordinates only.}
\label{image:block-diagram}
\end{figure}

We propose an extension of the projection-based approach \cite{Eichenseer2019} that projects the coordinates into different viewports in the perspective domain in order to realize differently oriented motion planes in 3D space. In this context, a viewport describes the orientation of a virtual perspective camera in 3D space. We allow the motion estimation procedure to choose between the front/back, the bottom/top, and the left/right viewport. We'll explain why it is possible to perform the calculations for opposite viewport pairs simultaneously, at the end of this section.

Fig. \ref{image:block-diagram} describes the general procedure of the viewport-adaptive motion compensation technique for one motion vector and viewport candidate. In a first step, the pixel coordinates of the current block $(x_\text{f}, y_\text{f}) \in \boldsymbol{B}_\text{cur}$ are projected to the unit sphere in 3D space. Spherical coordinates are used to describe the pixel coordinates $(\rho_\text{s}, \theta_\text{s}, \varphi_\text{s})$ on the unit sphere. As such, the radius is fixed to $\rho_\text{s} = 1$, $\theta_s$ is the incident angle obtained from the inverse fisheye projection, e.g.\ the inverse equisolid projection
\begin{align}
    \theta_\text{s} = 2 \arcsin\left(\frac{r_\text{f}}{2f}\right), 
\end{align}
and $\varphi_\text{s} = \varphi_\text{f}$ is equal to the angle of the pixel in polar coordinates $(r_\text{f}, \varphi_\text{f})$. In the following, we use the different coordinate systems interchangeably. Matching coordinates in the same context can be identified based on their subscripts. In a next step, the viewport rotation according to the desired virtual camera orientation is performed. This means that the pixel coordinates $(x_\text{s}, y_\text{s}, z_\text{s})$ on the unit sphere need to be rotated in 3D space using a suitable method for describing 3D rotations. As we only perform a rotation by $\pi/2$ around the $x$- and the $y$-axis, respectively, it is sufficient to define the rotations as simple transpositions of the 3D coordinates with
\begin{align}
    (x_\text{sr}, y_\text{sr}, z_\text{sr}) = (x_\text{s}, -z_\text{s}, y_\text{s})
\end{align}
for the bottom/top viewport and
\begin{align}
    (x_\text{sr}, y_\text{sr}, z_\text{sr}) = (z_\text{s}, y_\text{s}, -x_\text{s})
\end{align}
for the left/right viewport. For the front/back viewport, no rotation has to be performed. The rotated coordinates on the unit sphere are then projected to the perspective domain, obtaining $(r_\text{pr}, \varphi_\text{pr})$, where $r_\text{pr} = r_\text{p}(\theta_\text{sr})$ and $\varphi_\text{pr} = \varphi_\text{sr}$. The motion vector $\boldsymbol{m} = (\Delta x, \Delta y)$ is then applied to the viewport rotated coordinates, resulting in the viewport rotated, motion compensated coordinates
\begin{align}
    (x_\text{prm}, y_\text{prm}) = (x_\text{pr} + \Delta x, y_\text{pr} + \Delta y).
\end{align}
The coordinates $(x_\text{prm}, y_\text{prm})$ are further projected back onto the unit sphere using the inverse perspective projection
\begin{align}
    \theta_\text{srm} = \arctan\left(\frac{r_\text{prm}}{f}\right),
\end{align}
yielding $(\rho_\text{srm}, \theta_\text{srm}, \varphi_\text{srm})$ with $\rho_\text{srm} = 1$ and $\varphi_\text{srm} = \varphi_\text{prm}$. Afterwards, the viewport rotation is reversed according to
\begin{align}
    (x_\text{sm}, y_\text{sm}, z_\text{sm}) = (x_\text{srm}, z_\text{srm}, -y_\text{srm})
\end{align}
for the bottom/top viewport and
\begin{align}
    (x_\text{sm}, y_\text{sm}, z_\text{sm}) = (-z_\text{srm}, y_\text{srm}, x_\text{srm})
\end{align}
for the left/right viewport. Finally, the pixel coordinates are projected back to the fisheye domain, obtaining $(r_\text{fm}, \varphi_\text{fm})$, where $r_\text{fm} = r_\text{f}(\theta_\text{sm})$ and $\varphi_\text{fm} = \varphi_\text{sm}$. The resulting cartesian coordinates $(x_\text{fm}, y_\text{fm})$ are then used to extract the corresponding pixel values from the interpolated reference frame, yielding the motion compensated block $\boldsymbol{\hat{B}}_\text{cur}$ for the regarded motion vector and viewport candidate. Hence, besides the different motion vector candidates, the viewport-adaptive motion estimation procedure additionally tests the different viewports. The candidate yielding the minimum residual error (here: SSD) between the current block $\boldsymbol{B}_\text{cur}$ and the motion compensated block $\boldsymbol{\hat{B}}_\text{cur}$ is selected adaptively. The depicted procedure is repeated for all blocks in the current image to obtain the entire motion compensated frame $\boldsymbol{\hat{I}}_\text{cur}$. We want to emphasize that all calculations are solely based on the coordinates of the regarded block pixels and no projection of the image data into the perspective domain viewports is performed at any point.

\begin{figure}[t]
\centering
\includegraphics[width=0.91\columnwidth]{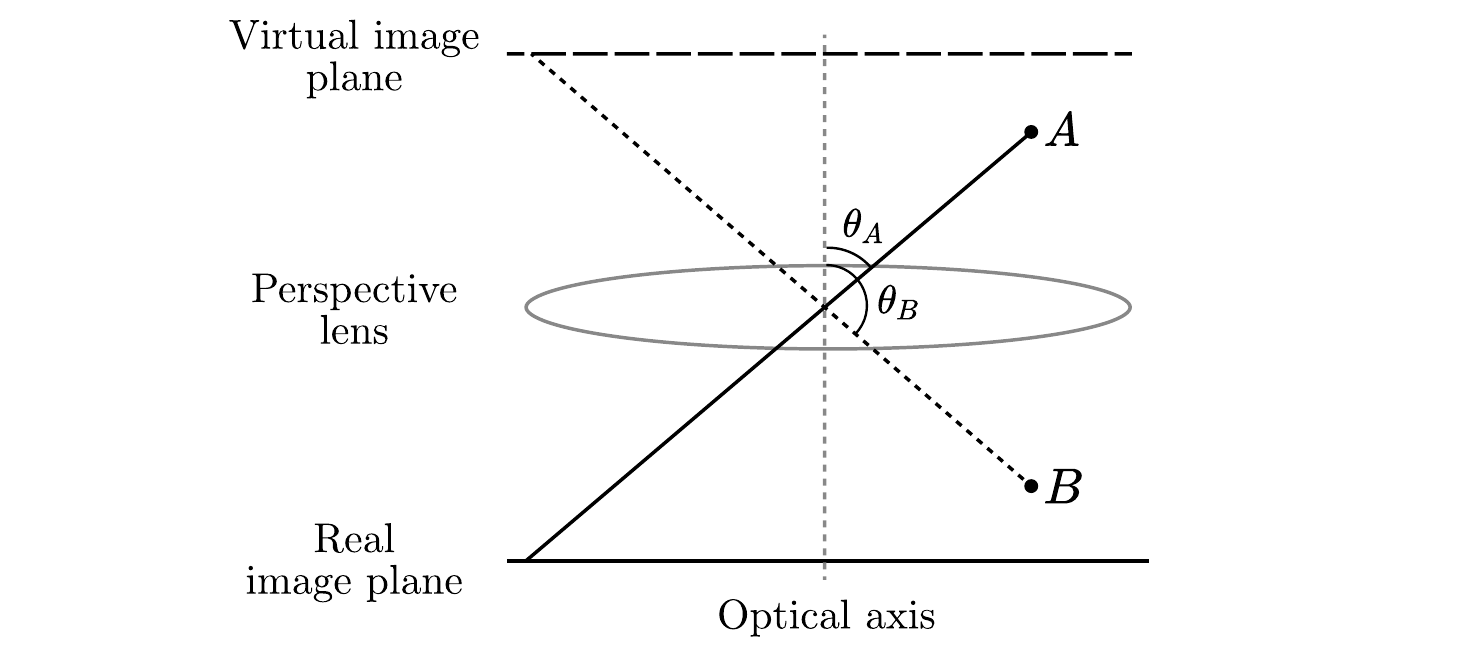}
\caption{Schematic of the perspective lens projection. Point $A$ with incident angle $\theta_A$ is projected onto the real image plane. Point $B$ with incident angle $\theta_B$ is projected onto the virtual image plane.}
\label{image:virtual-image-plane}
\end{figure}

As the perspective projection is only valid for incident angles $\theta < \pi/2$, special care has be taken regarding higher incident angles that occur in fisheye images. The basic problem is depicted in \mbox{Fig. \ref{image:virtual-image-plane}}, where the light ray with incident angle $\theta_A < \pi/2$ originating at point $A$ is mapped to the real image plane, while the light ray with incident angle $\theta_B > \pi/2$ originating at point $B$ is mapped to the virtual image plane. In the context of the performed projection of the viewport rotated pixel coordinates $(\rho_\text{sr}, \theta_\text{sr}, \varphi_\text{sr})$ to the perspective domain, any coordinates with $\theta_\text{sr} > \pi/2$ are therefore mapped to the virtual image plane. Regarding the investigated viewport pairs (e.g.\ front/back), the first viewport (e.g.\ front) always lies on the real image plane while the second viewport (e.g.\ back) lies on the virtual image plane. In \cite{Eichenseer2019}, a so-called ultra wide-angle compensation is performed to account for the virtual image plane. However, it does not provide exact reprojections into the spherical domain as it performs the compensation based on the reprojected radius, and hence, does not consider the nonlinearity of the projection functions. Therefore, we employ an advancement of the ultra wide-angle compensation called virtual image plane \mbox{compensation (VIPC)}, which conditionally performs three adjustments to the described viewport-adaptive procedure. If a pixel coordinate lies on the virtual image plane, (1) the motion vector  $\boldsymbol{m}$ for this coordinate is inverted, \mbox{(2) the} incident angle $\theta_\text{srm}$ received by the inverse perspective projection is corrected to $\theta_\text{srm}^\prime = \pi - \theta_\text{srm}$, and (3) the angle $\varphi_\text{srm}$ is corrected to $\varphi_\text{srm}^\prime = \varphi_\text{srm} - \pi$. By following this procedure for all pixel coordinates on the virtual image plane, exact reprojections are achieved and the calculations for both viewports in a viewport pair can be performed simultaneously.

\section{Simulations and Experiments}
\label{sec:experiments}

We investigate the performance of the proposed viewport-adaptive motion estimation technique by applying it to the task of motion compensation, which forms the baseline for many image and video processing tasks. In the following, three methods for block-based motion estimation and compensation employing a translatory motion model are tested. The traditional motion compensation method employing a translatory motion model is thereby called Translatory Motion Compensation (TMC) \cite{ShanZhu2000}, the projection-based approach is called Projection-based Translatory Motion Compen-\mbox{sation (PTMC) \cite{Eichenseer2019}}, and the novel viewport-adaptive approach is called Viewport-Adaptive Projection-based Translatory Motion Compensation (VA-PTMC).

All experiments are performed using the eight sequences \textit{HallwayA}, \textit{HallwayB}, \textit{HallwayD}, \textit{PillarsC}, \textit{PoolA}, \textit{PoolB}, \textit{PoolNightA} and \textit{Street} from the fisheye dataset \cite{Eichenseer2016}. The sequences of size 1088$\times$1088 pixels have been rendered using an equisolid fisheye lens with a bitdepth of 8 bits per channel and exhibit different global camera motion types such as translation, zoom and pan. For each sequence, 100 frame pairs consisting of a reference frame and the subsequent current frame are processed. The motion compensation is performed on the luminance channel with the Peak Signal to Noise Ratio (PSNR) and the Structural Similarity Index Measure (SSIM) \cite{Wang2004} as estimates for the achieved motion compensation quality. Both measurements are calculated on the actual pixel data only while the irrelevant area around the circular fisheye images is ignored.

A block matching search range of 96 pixels to all sides of the current block is selected for all algorithms. As the optimal brute-force full search strategy is rarely used in actual motion estimation systems due to its immense computational complexity, we employ the diamond search strategy \cite{ShanZhu2000}. This provides faster, more practical results and a better estimate of the resulting gains in quality for real applications. The sum of squared differences (SSD) is used as an error metric during the motion estimation procedure, and the motion vector candidates are limited to integer accuracy in the perspective domain. Different square blocksizes $B \times B$ with $B \in \{8, 16, 32, 64, 128\}$ are tested. For PTMC and VA-PTMC, the reference frame is interpolated to 1/8th pixel accuracy using cubic convolution interpolation \cite{Keys1981} and the motion compensated fisheye coordinates are quantized accordingly.

\begin{table}[t]
\footnotesize
\centering
\caption{Average PSNR results in dB of the motion compensated frames for the different sequences employing a blocksize of $B = 16$. The second column specifies the indices of the employed reference frames. The PSNR gain with respect to TMC is given in brackets.}\vspace{2pt}
\label{tab:psnr_results_sequences}
\begin{tabular}{lr|rrr}
\toprule
Sequence            & Frames   & TMC   &  PTMC         & VA-PTMC                \\
\midrule
\textit{HallwayA}   & 331--430 & 28.10 & 26.05 (--2.06) & \textbf{28.71 (+0.61)} \\
\textit{HallwayB}   &   1--100 & 31.12 & 29.68 (--1.45) & \textbf{33.21 (+2.09)} \\
\textit{HallwayD}   &   1--100 & 32.70 & 36.95 (+4.25) & \textbf{38.57 (+5.87)} \\
\textit{PillarsC}   & 101--200 & 36.78 & 37.43 (+0.65) & \textbf{38.52 (+1.74)} \\
\textit{PoolA}      & 101--200 & 36.09 & 37.40 (+1.31) & \textbf{39.42 (+3.33)} \\
\textit{PoolB}      & 301--400 & 36.22 & 38.20 (+1.98) & \textbf{41.17 (+4.95)} \\
\textit{PoolNightA} &   1--100 & 33.48 & 34.73 (+1.25) & \textbf{36.20 (+2.72)} \\
\textit{Street}     & 401--500 & 23.13 & 22.65 (--0.47) & \textbf{24.01 (+0.88)} \\
\midrule
Average             &          & 32.20 & 32.88 (+0.68) & \textbf{34.98 (+2.78)} \\
\bottomrule
\end{tabular}
\end{table}

Table \ref{tab:psnr_results_sequences} shows the achieved PSNR results for an exemplary blocksize of $B = 16$. It is clearly visible that the proposed VA-PTMC significantly outperforms TMC and PTMC. In the sequences \textit{HallwayA} and \textit{HallwayB}, that also exhibit non-translational motion, PTMC has difficulties and performs even worse than conventional TMC. VA-PTMC, on the other hand, provides considerable gains in quality for sequences exhibiting non-translational motion, as well. Furthermore, in sequence \textit{Street} large parts of the motion happen below the camera, i.e.\ on the bottom motion plane. The implicit front/back viewport of PTMC can not model the motion on that plane and hence leads to slightly worse results than TMC whereas VA-PTMC leads to a gain of +0.88 dB for the selected blocksize.

\begin{table}[t]
\footnotesize
\centering
\caption{Average PSNR results in dB for different blocksizes $B$. The PSNR gain with respect to TMC is given in brackets.}\vspace{2pt}
\label{tab:psnr_results_mean}
\begin{tabular}{l|rrr}
\toprule
Blocksize $B$ &   TMC &  PTMC &         VA-PTMC \\
\midrule
8             & 32.78 & 33.07 (+0.29) & \textbf{35.42 (+2.64)} \\
16            & 32.20 & 32.88 (+0.68) & \textbf{34.98 (+2.78)} \\
32            & 31.78 & 32.51 (+0.73) & \textbf{34.69 (+2.91)} \\
64            & 31.26 & 31.88 (+0.62) & \textbf{34.49 (+3.23)} \\
128           & 30.28 & 30.91 (+0.63) & \textbf{33.68 (+3.40)} \\
\midrule
Average       & 31.66 & 32.25 (+0.59) & \textbf{34.65 (+2.99)} \\
\bottomrule
\end{tabular}
\end{table}

Table \ref{tab:psnr_results_mean} shows the PSNR results for different blocksizes averaged over all sequences, i.e.\ averaged over 800 motion compensated frames. Clearly, the large gains of VA-PTMC with respect to TMC and PTMC are not limited to the investigated blocksize $B=16$ but are obtained for all blocksizes. While the overall motion compensation quality decreases for increasing blocksizes $B$, the gain of VA-PTMC with respect to TMC increases. Overall, VA-PTMC leads to average gains of +2.99 dB on the investigated fisheye sequences. For comparison, the average SSIM values amount to 0.8863 for TMC, 0.8924 (+0.0061) for PTMC, and 0.9227 (+0.0364) for VA-PTMC. With these significant improvements in terms of motion compensation quality, it can be expected that VA-PTMC leads to considerable improvements in algorithms that build on motion compensation, such as video coding, framerate up-conversion, or resolution enhancement.

\begin{figure}[t]
\centering
\includegraphics[width=0.9\columnwidth]{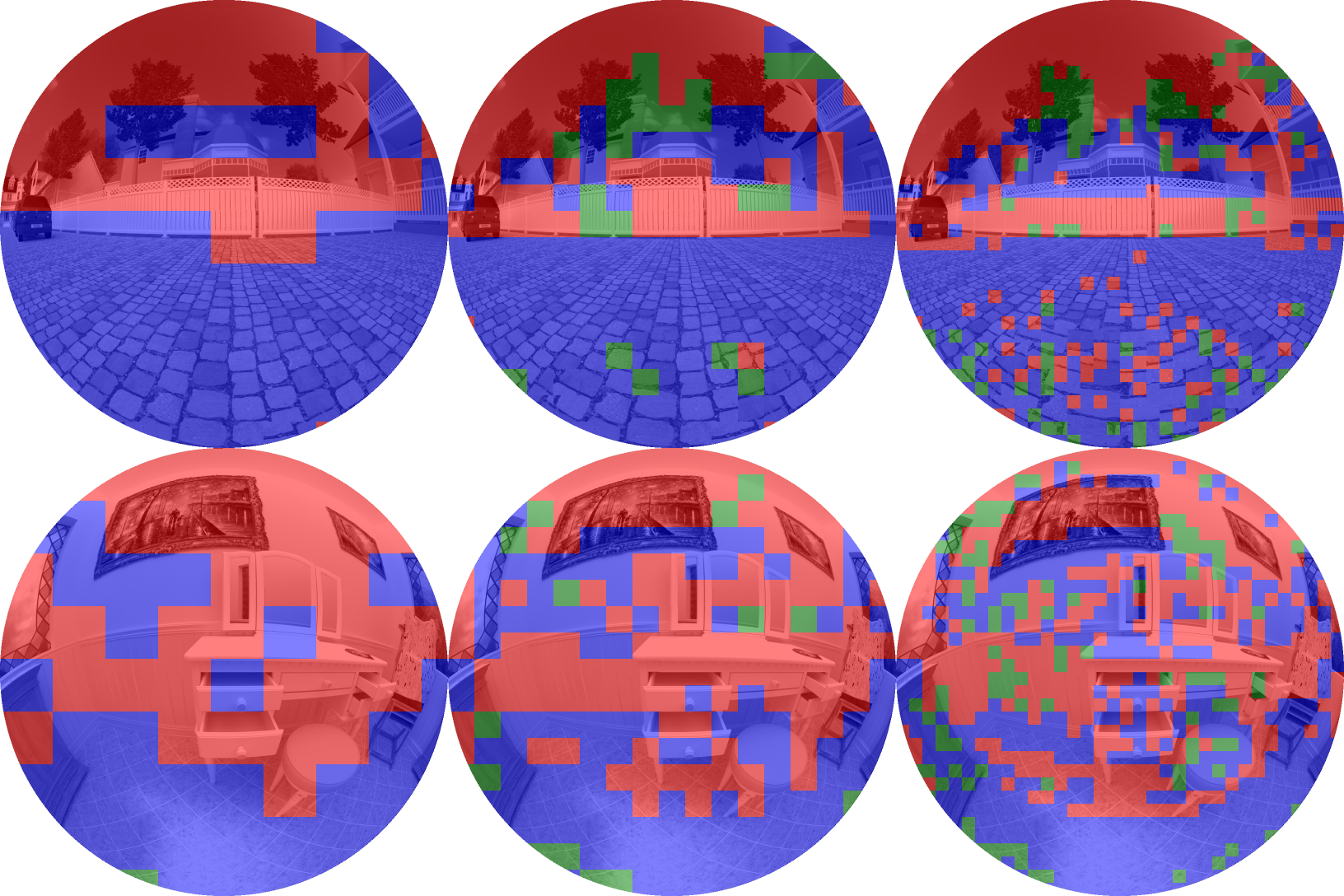}
\caption{Viewport decision maps overlaid on motion compensated frames of sequences \textit{Street} (top) and \textit{HallwayD} (bottom) for blocksizes 128, 64, and 32 (from left to right). The viewports are color coded as front/back (red), bottom/top (blue), left/right (green).}
\label{image:decision-maps}
\end{figure}

Fig. \ref{image:decision-maps} shows the color coded viewport decisions overlaid on the motion compensated frames for the sequences \textit{Street} and \textit{HallwayD} for the blocksizes $B = 128$, $B = 64$ and $B = 32$. It is visible that the proposed VA-PTMC reliably selects the bottom/top viewport for structured ground areas in the images, especially for larger blocksizes. For smaller blocksizes, noise-like patterns emerge in some image areas, where viewports that do not match the expected motion plane supplied the minimum SSD. This can be partly attributed to the suboptimal diamond search strategy. It has shown that in some cases, unexpected viewport decisions might be beneficial in order to implicitly access pixels at subpixel precision. Depending on the application scenario, this behavior might be desired. In other cases, a possible method to prevent these situations could be a voting system that takes different viewport decisions within a suitable neighborhood into account and then selects the most frequent viewport for the current block.

\begin{figure}[t]
\setlength\figurewidth{\columnwidth}
\setlength\figureheight{.62\columnwidth}
\centering
{\footnotesize\input{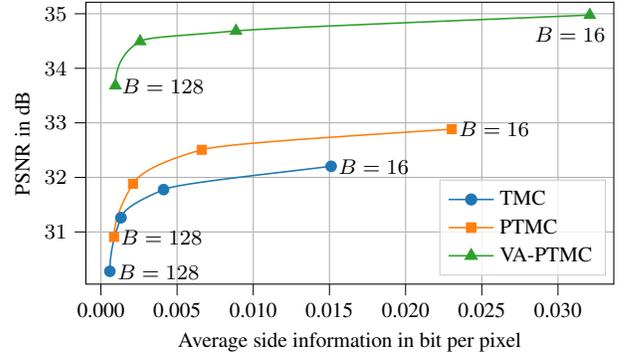}}
\caption{Motion compensation PSNR over average side-information in bit per pixel. For TMC and PTMC, the side information consists of the motion vectors only, while for VA-PTMC the viewport decision is included additionally.}
\label{fig:rate_distortion}
\end{figure}

Fig. \ref{fig:rate_distortion} shows the average compressed side information per pixel in relation to the average PSNR of the motion compensated images for different blocksizes $B$. Thereby, the side information is compressed using the bzip2 algorithm \cite{Seward2019}. For TMC and PTMC, the side information consists of the 8 bit signed integer motion vectors only, while for VA-PTMC the 2 bit viewport decision is included additionally. The more accurate fisheye motion model provides slighlty better results than TMC at comparable rates. VA-PTMC, however, shows significantly improved results and spans roughly the same range as TMC and PTMC in terms of side information rate. Especially noteworthy is that neither PTMC nor TMC can match the high quality of VA-PTMC for $B=128$ with any of the tested blocksizes. This strongly suggests that the additional bits pay off and notable improvements in rate-distortion performance can be expected if the proposed viewport-adaptive approach is to be integrated into an actual video codec.

\vspace*{-8.75pt}
\section{Conclusion}
\label{sec:conclusion}

In this paper, we proposed a novel viewport-adaptive motion estimation technique for fisheye images that leads to considerable gains over the projection-based approach and outperforms traditional motion compensation by almost 3 dB in terms of PSNR on fisheye images. We have shown that our method leads to considerable gains even on sequences where the prior projection-based approach failed. Hence, significant gains for other algorithms based on motion compensation can be expected. An investigation of the applicability to video coding suggests that major improvements in rate-distortion performance are highly likely. As a next step, we will therefore integrate our method into the VVC Test Model (VTM) \cite{Chen2020}, which will require additional research to adapt other dependent algorithms in the coding pipeline to the presented viewport-adaptive approach.

\newpage
\balance
\bibliographystyle{IEEEbib}
\bibliography{ms}

\end{document}